\documentclass{article}

\usepackage[final,nonatbib]{nips_2017}

\usepackage[utf8]{inputenc} 
\usepackage[T1]{fontenc}    
\usepackage{hyperref}       
\usepackage{url}            
\usepackage{booktabs}       
\usepackage{amsfonts}       
\usepackage{microtype}      
\usepackage{subfigure}
\usepackage{epsfig}
\usepackage{graphicx}
\usepackage{amsmath}
\usepackage{amssymb}
\usepackage{subfigure}
\usepackage{float}

\title{SquishedNets: Squishing SqueezeNet further for edge device scenarios via deep evolutionary synthesis}

\author{
  Mohammad Javad Shafiee\\
  Dept. of Systems Design Engineering\\
  University of Waterloo, DarwinAI\\
  \texttt{mjshafiee@uwaterloo.ca}
\And
  Francis Li\\
  Dept. of Systems Design Engineering\\
  University of Waterloo, DarwinAI\\
  \texttt{francis@darwinai.ca}
  \And
  Brendan Chwyl\\
  Dept. of Systems Design Engineering\\
  University of Waterloo, DarwinAI\\
  \texttt{brendan@darwinai.ca}
\And
  Alexander Wong\\
  Dept. of Systems Design Engineering\\
  University of Waterloo, DarwinAI\\
  \texttt{a28wong@uwaterloo.ca} \\
}

\begin{document}

\maketitle

\begin{abstract}
While deep neural networks have been shown in recent years to outperform other machine learning methods in a wide range of applications, one of the biggest challenges with enabling deep neural networks for widespread deployment on edge devices such as mobile and other consumer devices is high computational and memory requirements.  Recently, there has been greater exploration into small deep neural network architectures that are more suitable for edge devices, with one of the most popular architectures being SqueezeNet, with an incredibly small model size of 4.8MB.  Taking further advantage of the notion that many applications of machine learning on edge devices are often characterized by a low number of target classes, this study explores the utility of combining architectural modifications and an evolutionary synthesis strategy for synthesizing even smaller deep neural architectures based on the more recent SqueezeNet v1.1 macroarchitecture (considered state-of-the-art in efficient architectures) for applications with fewer target classes.  In particular, architectural modifications are first made to SqueezeNet v1.1 to accommodate for a 10-class ImageNet-10 dataset, and then an evolutionary synthesis strategy is leveraged to synthesize more efficient deep neural networks based on this modified macroarchitecture.  The resulting \textbf{SquishedNets} possess model sizes ranging from 2.4MB to 0.95MB ($\sim$5.17X smaller than SqueezeNet v1.1, or 253X smaller than AlexNet). Furthermore, the SquishedNets are still able to achieve accuracies ranging from 81.2\% to 77\%, and able to process at speeds of 156 images/sec to as much as 256 images/sec on a Nvidia Jetson TX1 embedded chip.  These preliminary results show that a combination of architectural modifications and an evolutionary synthesis strategy can be a useful tool for producing very small deep neural network architectures that are well-suited for edge device scenarios without the need for compression or quantization.
\end{abstract}

\section{Introduction}

One can consider deep neural networks~\cite{lecun2015deep} to be one of the most successful machine learning methods, outperforming many state-of-the-art machine learning methods in a wide range of applications ranging from image categorization~\cite{krizhevsky2012imagenet} to speech recognition.  A very major factor to the great recent successes of deep neural networks has been the availability of very powerful high performance computing systems due to the great advances in parallel computing hardware.  This has enabled researchers to significantly increase the depth and complexity of deep neural networks, resulting in greatly increased modeling capabilities.  As such, the majority of research in deep neural networks have largely focused on designing deeper and more complex deep neural network architectures for improved accuracy.  However, the increasing complexity of deep neural networks has become one of the biggest obstacles to the widespread deployment of deep neural networks on edge devices such as mobile and other consumer devices, where computational, memory, and power capacity is significantly lower than that in high performance computing systems.

Given the proliferation of edge devices and the increasing demand for machine learning applications in such devices, there has been an increasing amount of research exploration on the design of smaller, more efficient deep neural network architectures that can both infer and train faster, as well as transfer faster onto embedded chips that power such edge devices.  One commonly employed approach for designing smaller neural network architectures is synaptic precision reduction, where the number of bits used to represent synaptic strength is significantly reduced from 32-bit floating point precision to fixed-point precision~\cite{Shin2017}, 2-bit precision~\cite{li2016ternary,Yin2017,Meng2017}, or 1-bit precision~\cite{Courbariaux2015,Rastegari2015}.  While this approach leads to greatly reduced model sizes, the resulting deep neural networks often require specialized hardware support for accelerated deep inference and training on embedded devices, which can limit their utility for wide range of applications.

Another approach to designing smaller deep neural network architectures is to take a principled approach and employ architectural design strategies to achieve more efficient deep neural network macroarchitectures~\cite{MobileNet,SqueezeNet}.  An exemplary case of what can be achieved using such an approach is SqueezeNet~\cite{SqueezeNet}, where three key design strategies where employed: 1) decrease the number of 3x3 filters, 2) decrease the number of input channels to 3x3 filters, and 3) downsample late in the network.  This resulted in a macroarchitecture composed of Fire modules that possessed an incredibly small model size of 4.8MB, which is 50X smaller than AlexNet with comparable accuracy on ImageNet for 1000 classes.  The authors further introduced SqueezeNet v1.1, where the number of filters as well as the filter sizes are further reduced, resulting in 2.4X less computation than the original SqueezeNet without sacrificing accuracy and thus can be considered state-of-the-art in efficient network architectures.

Inspired by the incredibly small macroarchitecture of SqueezeNet, we are motivated to take it one step further by taking into account that the majority of applications of machine learning on edge devices such as mobile and consumer devices are quite specialized and often require a much fewer number of target classes (typically less than 10 target classes).  As such, this study explores the utility of combining architectural modifications and an evolutionary synthesis strategy for synthesizing even smaller deep neural architectures based on the SqueezeNet v1.1 macroarchitecture for applications with fewer target classes.  We will refer to these smaller deep neural network architectures as \textbf{SquishedNets}.

\vspace{-0.2cm }
\section{Architectural modification for fewer target classes}
\vspace{-0.1 cm}

The first and simplest strategy taken in this study is to perform some simple architectural modifications to the macroarchitecture of SqueezeNet v1.1 to accommodate for scenarios where we are dealing with much fewer target classes.  For this study, we explored a classification scenario where the number of target classes is reduced to 10, as fewer target classes is quite common for many machine learning applications on edge devices such as mobile and other consumer devices.  Given the reduced number of target classes, we modify the \textbf{conv10} layer to a set of 10 1x1 filters.  Given that the \textbf{conv10} layer contains $\sim$40\% of all parameters in SqueezeNet~\cite{SqueezeNet}, this architectural modification resulted in a significant reduction in model size.  While not a leap of the imagination and a rather trivial modification, this illustrates that SqueezeNet v1.1 is a great macroarchitecture that can be modified to be even more efficient for edge scenarios where there are fewer target classes.

\vspace{-0.2cm }
\section{Evolutionary synthesis of more efficient network architectures}
\vspace{-0.1 cm}

The second strategy taken in this study is to employ an evolutionary synthesis strategy~\cite{javad2016evonet,shafiee2016evolutionary,shafiee2016evolutionary2} to synthesize even more efficient deep neural network architectures than can be achieved through principled macroarchitecture design strategies.  First proposed by~\cite{javad2016evonet} and subsequently extended~\cite{shafiee2016evolutionary,shafiee2016evolutionary2}, the idea behind evolutionary synthesis is to automatically synthesize progressively more efficient deep neural networks over successive generations in a stochastic synaptogenesis manner within a probabilistic framework.  More specifically, synaptic probability models are used to encode the genetic information of a deep neural network, thus mimic the notion of heredity.   Offspring networks are then synthesized in a stochastic manner given the synaptic probability models and a set of computational environmental constraints for influencing synaptogenesis, thus forming the next generation of deep neural networks, thus mimicking the notions of random mutation and natural selection.  The offspring deep neural networks are then trained and this evolution synthesis process is performed over generations until the desired traits are met.

While a more detailed description of the evolutionary synthesis strategy can be found in~\cite{javad2016evonet}, a brief mathematical description is provided as follows.  Let the genetic encoding of a network be formulated as $P(H_g|H_{g-1})$, where the network architecture $H_g$ at generation $g$ is influenced by the network architecture $H_{g-1}$ of generation \mbox{$g-1$}.  An offspring network is synthesized in a stochastic manner via a synthesis probability $P(H_g)$, which combines the genetic encoding $P(H_g|H_{g-1})$ with the environmental factor model $R$ being imposed:
\begin{align}
P(H_g) \approx P(H_g|H_{g-1}) \cdot R.
\end{align}
Since the underlying goal is to influence the synaptogenesis behaviour of offspring deep neural networks to be progressively more efficient generation after generation, the environmental factor model $R$ is set to a value less than 1 to further enforce a resource-starved environment on the deep neural networks from generation to generation.  In this study, the aforementioned modified  macroarchitecture is used as the ancestral precursor for the evolutionary synthesis process to produce deep neural networks with even smaller network architectures.  In this study, 15 generations of evolutionary synthesis was performed to produce the final SquishedNets.

\begin{table}
\begin{center}
	\footnotesize
	\caption{ImageNet-10 dataset}
	\label{Tab:Class}
		\begin{tabular}{|c|c||c|c|}\hline
			wnid & Class Name & wnid & Class Name \\ \hline \hline
			n02783161&  pen & n03584254 &cell phone\\
			n03085013& keyboard &n04548362& wallet\\
		n04557648& water bottle& n07930864 &cup\\
		n04037443& car&  n03782006& monitor\\
		n03793489 &computer mouse &n04409515& tennis ball \\\hline
		\end{tabular}
	\end{center}	
\end{table}

\begin{table}[ht]
	\begin{center}
		\footnotesize
		\caption{Performance results of SquishedNets. }
		\label{Tab:CPU}
		\begin{tabular}{|c|c||c|c|c|c|}
			\hline
			Model  & Model & Reduction in & Reduction in & Runtime  & Top-1  \\
			  Name & size & model size & model size &  speed  & accuracy   \\
			    & & vs. SqueezeNet v1.1 & vs. AlexNet & (images/sec) & (ImageNet-10)   \\\hline \hline
			  SquishedNet-1&2.4MB & 2.04X &100X &  156.09& 81.2\% \\
			  SquishedNet-2&2.0MB & 2.45X & 120X & 174.86 & 79.6\% \\
			 SquishedNet-3&1.3MB & 3.77X & 184X & 225.35 & 78.6\% \\
			 SquishedNet-4&0.95MB & 5.17X & 253X &256.00  & 77.0\% \\\hline
		\end{tabular}
	\end{center}
\end{table}

\vspace{-0.2 cm}
\section{Preliminary Results and Discussion}
\vspace{-0.1 cm}
To study the utility of a combination of architectural modifications and evolutionary synthesis on synthesizing very small deep neural network architectures based on the SqueezeNet v1.1 macroarchitecture that are well-suited for edge device scenarios, we examine the top-1 accuracies and runtime speeds (on an Nvidia Jetson TX1 embedded chip with batch size of 32) of our synthesized \textbf{SquishedNets} on the 10-class ImageNet-10 dataset.  The ImageNet-10 dataset used in this study is a subset of the ImageNet dataset composed of the following ten target classes reported in Table~1.

The performance results of four different SquishedNets (produced at four different generations of the evolutionary synthesis process) are shown in Table~1.  A number of observations can be made based on Table~2.  First, it can be observed that leveraging both architectural modifications to account for fewer target classes as well as evolutionary synthesis results in the generation of even more efficient network architectures, as evident by the SquishedNets having model sizes range from 2.4MB to just 0.95MB.  Therefore, the smallest SquishedNet is 5.17X smaller than SqueezeNet v1.1 (or $\sim$253X smaller than AlexNet).  Second, not only was there a significant model size reductions, the SquishedNets were able to process at speeds of 156 images/sec to as much as 256 images/sec on a Nvidia Jetson TX1 embedded chip, which is significant particularly for edge scenarios with mobile and other consumer devices.  Therefore, the ability to not only achieve very small model sizes but also fast runtime speeds has great benefits when used in resource-starved environments with limited computational, memory, and energy requirements.  In terms of top-1 accuracy, the SquishedNets are able to still achieve accuracies ranging from 81.2\% to 77.0\%, which is high enough for many edge applications. These preliminary results show that a combination of architectural modifications and an evolutionary synthesis strategy can be a useful tool for producing very small deep neural network architectures that are well-suited for edge device scenarios without the need for compression or quantization.

\vspace{-0.35 cm}
\section*{Acknowledgment}
\vspace{-0.15 cm}
The authors thank NSERC, the Canada Research Chairs program, Nvidia, and DarwinAI.

{\small
\bibliographystyle{plain}
\bibliography{ccn_style}
}

\end{document}